# To use or not to use proprietary street view images in (health and place) research? That is the question

Marco Helbich[1,*,#], Matthew Danish[1,#], SM Labib[1], Britta Ricker[2]

[1] Department of Human Geography and Spatial Planning, Utrecht University

[2] Copernicus Institute of Sustainable Development, Utrecht University

[*] Corresponding author: Department of Human Geography and Spatial Planning. Faculty of Geosciences, Utrecht University, Princetonlaan 8a, 3584 CB, Utrecht, The Netherlands, www: http://www.uu.nl/staff/mhelbich, Tel.: (0031) 30 253 2017, Email: m.helbich@uu.nl

[#] Both authors contributed equally.

## Abstract

Computer vision-based analysis of street view imagery has transformative impacts on environmental assessments. Interactive web services, particularly Google Street View, play an ever-important role in making imagery data ubiquitous. Despite the technical ease of harnessing millions of Google Street View images, this article questions the current practices in using this proprietary data source from a European viewpoint. Our concern lies with Google's terms of service, which restrict bulk image downloads and the generation of street view image-based indices. To reconcile the challenge of advancing society through groundbreaking research while maintaining data license agreements and legal integrity, we believe it is crucial to 1) include an author's statement on using proprietary street view data and the directives it entails, 2) negotiate academic-specific license to democratize Google Street View data access, and 3) adhere to open data principles and utilize open image sources for future research.

# Street view imagery as a curse or blessing

The advent of street view imagery has revolutionized how the environment is assessed and sparked interest in health and place research (Rzotkiewicz et al. 2018). Street view images offer an alternative perspective on places, capturing the environment through the eyes of people on the ground rather than the conventional viewpoint from space through Earth observation (Helbich et al. 2021). Consequently, this wealth of data helps answer novel research questions dealing with streetscapes that would be intractable otherwise (see, e.g., Biljecki & Ito 2021 for an overview). However, the providers of this street view imagery impose contractual terms that raise legal concerns, particularly in the European Union (EU). In this article, we share and reflect upon valuable insights from ongoing discussions between our research group and a representative of legal affairs at Utrecht University, as they are relevant to the broader community making use of street view images in research.

The usefulness and soundness of street view imagery as a large-scale data source for environmental studies usually requires hundreds of thousands of images (or more), which makes self-collection of imagery impractical due to temporal and monetary constraints that apply to most academic research. Consequently, researchers typically depend on street view images sourced from others, whether proprietary or volunteered. Most studies published over the last decade have relied on proprietary imagery (Biljecki & Ito 2021), which is abundant and readily available through programmable interfaces.

Google is the most popular supplier of such proprietary street view imagery, particularly in the Global North, through the Google Maps web service. This mapping service enables users to virtually navigate urban landscapes based on Google Street View, operating since 2007 in over 80 countries. Google Street View imagery comes with several advantages, including consistent data acquisition through high-end cameras mounted on vehicles, yielding homogeneous data quality of high-resolution images and 360-degree panoramic views of streets typically captured along public roads (Anguelov et al. 2010).

While manual image interpretation or semi-automated information extraction of these street view images for location-based studies through image data processing software is cumbersome, advancements in computer vision have been a game-changer in time and effort saved. Deep learning algorithms typically run on thousands, if not millions, of web-scraped images to identify hundreds of different objects visible in these images (e.g., trees, roads, and bicycles). Harnessing millions of Google

Street View images feeding these models is technically straightforward via Google's Application Programming Interface, which suggests endless research opportunities (Biljecki & Ito 2021), particularly in dealing with the assessment of the neighborhood environment including green space, safety issues, walkability, and more. This could be good news; however, the current terms of service for using the Google Maps platform impose conditions that render the use of its imagery legally questionable as a data source for researchers.

### Google's terms of service and legal viewpoints

The Google Maps Platform Terms of Service lay out the possibilities and restrictions on Google's intellectual property for each aspect of Google Maps, including Google Street View, even for non-commercial use (Google 2024a). This includes any actions of bulk-downloading and analyzing imagery obtained from Google, which fall under the general category of text and data mining research. In the United States, it may be possible to argue that text and data mining is permitted under the Fair Use Doctrine (Copyright Act of 1973), and some text and data mining cases have held up in court, although it is not clear how much this can be relied upon (Kollár 2021). Rundle et al. (2022) noted that invoking fair usage over Google's term of service is an area of active litigation and is currently unsettled.

In any case, such doctrine is specific to United States law and does not exist in the EU. The authors of this article are EU-based and, therefore, take a particular interest in this context. The EU has recently (2019) created new rules about text and data mining that could potentially be useful for researchers who wish to legally analyze Google Street View images. Specifically, Directive (EU) 2019/790 Title II Article 3 (European Union 2019) (from here on known simply as 'Article 3') allows "*for reproductions and extractions made by research organisations and cultural heritage institutions in order to carry out, for the purposes of scientific research, text and data mining of works or other subject matter to which they have lawful access*". Each EU member state must enact its own laws in compliance with this directive; in the case of the Netherlands, for example, it was implemented by the Dutch Copyright Act, article 15n, that came into effect on 7 July 2021, and which largely follows the wording of the EU directive, albeit translated into Dutch (Staatsblad 2020).

However, the major components of Article 3 are vaguely defined and could be unfavorably interpreted by courts in the case of legal action taken against researchers (Kollár 2021). In particular, 'reproductions and extractions' may not be sufficient to cover the activities of street view image surveying and processing, 'purposes of scientific research' are not clearly defined, and 'lawful access' could be construed against researchers who bulk-download Google Street View images in contravention of the terms of service.

In brief, the Google Maps Platform Terms of Service (Google 2024a) outlines in §3.2.3 (Restrictions Against Misusing the Services) that, first, "*customer will not export, extract, or otherwise scrape Google Maps Content*", and will not "*pre-fetch, index, store, reshare, or rehost*" or "*bulk download […] Street View images*". Second, it states that customers "*will not create content based on Google Maps Content*"; particularly emphasized is that customers are not allowed to ''*construct an index of tree locations within a city from Street View imagery*". Finally, and most important, the Geo Guidelines (Google 2024b) also indicate that academic use is no exception ("*…restrictions apply to all academic, nonprofit, and commercial projects*") and that no exceptions are granted. Although these conditions apply to the vast majority of studies, the possibility of additional contracts with Google exists, providing specific exceptions to these general terms of service. However, we do not consider those because they require special negotiations that would exclude most academic researchers.

We break down the concrete steps that form a pipeline for research using street view imagery and examine how each one may contravene the Google terms of service or not be permitted even under the regime of Article 3, bearing in mind that the authors of this article are not lawyers and nothing written in this document can be taken as legal advice:

1. The available street view images are systematically queried (e.g., every 10 m along the roads) and downloaded for an area under investigation. This requires scraping, storing, and bulk-downloading content, which Google does not permit. It is unclear whether images obtained this way are considered 'lawfully accessed' for use in data mining under Article 3.
2. The images are preprocessed (e.g., cropped to the same dimension) and fed into a database. This is a form of pre-fetching, storing, and indexing and, therefore, may not be permitted by Google. This probably falls under older 'temporary copy' exceptions to EU copyright rules (see Directive 2001/29/EC), especially if this database is not shared with anyone else.
3. Pre-trained deep learning algorithms are applied to extract image content (e.g., identifying which pixels correspond to portions of vegetation, roads, cars, and buildings). This would run afoul of the indexing and content creation clauses in the terms of service. However, Article 3 would override those restrictions if the images are lawfully accessed and used only for research purposes.
4. The processed images may be shown to participants in a scientific survey. This could be considered resharing and rehosting, which are not permitted as per Google's terms of use. It is unclear if anything in EU copyright law would protect researchers in this case.
5. The images and the results of the survey may be collected into a large dataset. This falls under Google's restrictions against storing, content creation, and indexing, and may also be resharing and rehosting if the dataset is published. Article 3 will probably not apply if the images are reshared. However, it may apply if only the survey results are reshared (and the images are omitted, possibly replaced by references such as URLs directing views to the corresponding location within the Google Street View platform).
6. This dataset (from point 5) may be used to train additional deep learning models that are then leveraged for future research. This can be construed as indexing and content creation, disallowed by Google. Article 3 can likely be used to override the Google terms of service provided that the dataset complies with the aforementioned lawful access requirement and that the purpose of these models is for scientific research. It is unclear whether further uses of these generated models must also be scientific research and how that can possibly be enforced.

Therefore, even with the exemptions granted under legislation enacted to fulfill Article 3, it remains questionable whether proprietary street view images from Google are available for research purposes. Even if they prevail, the possibility of legal action has a chilling effect on the accessibility and reusability of the data. To date, there has been no jurisprudence on any of the legislation enacted under Article 3. Until a judge issues a ruling based on a comparable case, preferably at the EU-level so the precedent has EU-wide applicability, the use of Google Street View images carries the potential risk of legal action against researchers. In the longer term, a solution such as that proposed by Kollár (2021), using ideas from Japanese copyright law, could provide the desired legal certainty for researchers to work soundly using proprietary street view imagery; however, that kind of regulatory change does not appear to be coming forth in the foreseeable future.

### A call to action for research practices

We see three ways forward for the time being: First, to relieve the burden on editors and safeguard them against possible future legal ramifications that published research is based on proprietary data, it remains the authors' obligation to submit legally sound research, warranting an explicit explanation of terms of use and any legal ground they must use such data as part of the manuscript submission process. Notably, authors should explicitly state in the manuscript (either in the Methods section, or in the Acknowledgements) that the data were lawfully accessed (e.g., fees were paid), and whether they have invoked their country's legal framework for text and data mining (e.g., fair use in the United States or an Article 3 implementation in the EU) in collecting and analyzing the images used in the study. In cases where legal implications are possibly involved, editors (and publishers) may seek advice from legal professionals or the publisher's legal counsel to understand the potential legal ramifications and decide on an appropriate course of action.

Second, although software and data often come with various licensing options, the same does not apply to Google Street View imagery. While occasional 'pay to play' arrangements may be in place between Google and select research groups negotiated on a case-by-case basis, we advocate for Google (and other proprietary providers) to release their imagery under a license designed to permit non-commercial academic use. Such an academic license would democratize Google Street View data access in academia, promoting equity and inclusivity in research, and would help simplify legal compliance.

Third, in the spirit of open and reproducible science, we call on researchers to use street view image sources that comply with open data (Singleton et al. 2016) and the Findability, Accessibility, Interoperability and Reusability (FAIR) principles for scientific data management (Wilkinson et al. 2016) instead of proprietary sources. At this time, a handful of municipalities supply their own self-collected street view data repositories, ensuring data uniformity across their specific geography (e.g., Amsterdam, The Netherlands). At a larger scale, one of the most substantial sources of FAIR-compliant street view images is Mapillary (owned by Meta), which offers volunteered/crowd-source images that are freely available under Creative Commons Share-Alike With-Attribution terms from a well-documented public application programming interface (Alvarez Leon, & Quinn 2019). Prospective researchers must take some preprocessing steps to use imagery from sources that are from volunteers. For example, Mapillary relies extensively on user-submitted imagery, and the quality of such imagery can vary considerably, from unusable low-quality, low-resolution photographs with poor lighting or other problems, all the way up to professional-level, high-quality panoramic images competitive with proprietary sources.

Data processing pipelines to address these challenges are on the way (Danish et al. 2024). For example, Zheng and Amemiya (2023), as well as Sánchez and Labib (2024), showed that such user-submitted street view images can be usefully filtered, selected, and preprocessed to produce a usable dataset of suitable imagery. This is additional work; however, the reward is that projects using FAIR image sources such as Mapillary can offer reproducible processes that comply with legal terms of use and can be used as building blocks for further research.

Conclusions

Taken together, in this article, we caution against the uncritical use of proprietary street view imagery for research purposes, due to possible legal issues which are often overlooked. Despite Google Street View being an extremely useful dataset, the existing terms of service preclude the use of their imagery for many research purposes, and the continued use of this imagery may pose some legal risks to researchers and institutions in the future. We see this as a key dilemma for researchers eager to conduct cutting-edge research while adhering to the law and producing legally reproducible methods and results.

While our primary intention is to reflect on the critical use of proprietary street view data for research purposes, we would like to emphasize that we seek neither to discredit previous studies nor to question their ethical practices in some way. We recognize that our interpretation of the terms of service and relevant law may not correspond with other people's interpretation; we are not lawyers and this document is not legal advice. Our concern lies mainly with the legal framework within the EU because that is where we are based; however, researchers located in other jurisdictions may have different issues and perspectives. The EU has introduced a copyright directive (Article 3) intended to enable academic use of data mining techniques; however, all implementations of this directive are recent and the resulting laws have yet to be tested in court. Therefore, EU-based researchers relying on a legal interpretation of their country's corresponding implementation are taking a legal risk that could result in litigation against them. Furthermore, even if judges take a charitable view of Article 3, there are some research purposes which may not fall within its protection at all (e.g., re-publishing images as part of a dataset).

Regardless of whether Google Street View turns out to be usable for some or all researchers, we believe that these concerns about legal issues surrounding proprietary street view images should give researchers pause to critically judge legal concerns of using such sources. We advocate for three measures concerning future research practices:

- First, we suggest implementing an author's declaration regarding the utilization of proprietary street view data and the directives it entails.
- Second, to democratize Google Street View data access, we urge for a license designed for academic purposes.
- Third, we call for those who find open and reproducible science important to promote and make use only of open and FAIR-compliant street view images, and to campaign for more such open datasets, to ensure the integrity of work built on these datasets.

## Acknowledgments

The content of our commentary was discussed with and substantiated by Jan van de Pas, legal affairs at Utrecht University. We appreciate Jan's insights. Further, we express our gratitude to the reviewers for their constructive comments that have enhanced the quality of the article.

## Conflict of interest

MH is an associate editor of Health and Place but took no part in the peer review and decision-making processes for this article. The other authors have no conflict of interest to declare.

## CRediT author statement

MH: Writing - Original Draft; MD: Writing - Original Draft; SML: Writing - Review & Editing; BR: Writing - Review & Editing

## Funding

This work received funding from the Faculty of Geosciences, Utrecht University (BMD 3.1fb220215).